\documentclass[runningheads]{llncs}
\usepackage{graphicx}
\usepackage{comment}
\usepackage{amsmath,amssymb} 
\usepackage{color}
\usepackage{adjustbox}
\usepackage{overpic}
\usepackage{booktabs,siunitx}
\usepackage{siunitx}
\usepackage{subfigure}
\usepackage{caption}
\usepackage{sidecap}
\usepackage{algorithm}
\usepackage{algorithmic}
\usepackage{fdsymbol}
\newcommand{\m}{Moir{\'e}~}
\newcommand{\mt}{Moir{\'e}Tag~}
\newcommand{\na}{n_0}
\newcommand{\nb}{n_1}
\newcommand{\vx}{x}
\newcommand{\vd}{d}

\begin{document}

\title{QR-Tag: Angular Measurement and Tracking with a QR-Design Marker } 

\titlerunning{QR-Tag}
%
\author{Simeng Qiu \and Hadi Amata \and Wolfgang Heidrich}
\authorrunning{S. Qiu et al.}
\institute{King Abdullah University of Science and Technology \\ \email{simeng.qiu@kaust.edu.sa}}

\maketitle
\begin{abstract}
Directional information measurement has many applications in domains such as robotics, virtual and augmented reality, and industrial computer vision. Conventional methods either require pre-calibration or necessitate controlled environments. The state-of-the-art \mt approach exploits the \m effect and QR-design to continuously track the angular shift precisely. However, it is still not a fully QR code design. To overcome the above challenges, we propose a novel snapshot method for discrete angular measurement and tracking with scannable QR-design patterns that are generated by binary structures printed on both sides of a glass plate. The QR codes, resulting from the parallax effect due to the geometry alignment between two layers, can be readily measured as angular information using a phone camera. The simulation results show that the proposed non-contact object tracking framework is computationally efficient with high accuracy. 
\end{abstract}
\vspace{-20pt}
\begin{figure}
	\centering
	\includegraphics[width=.75\linewidth]{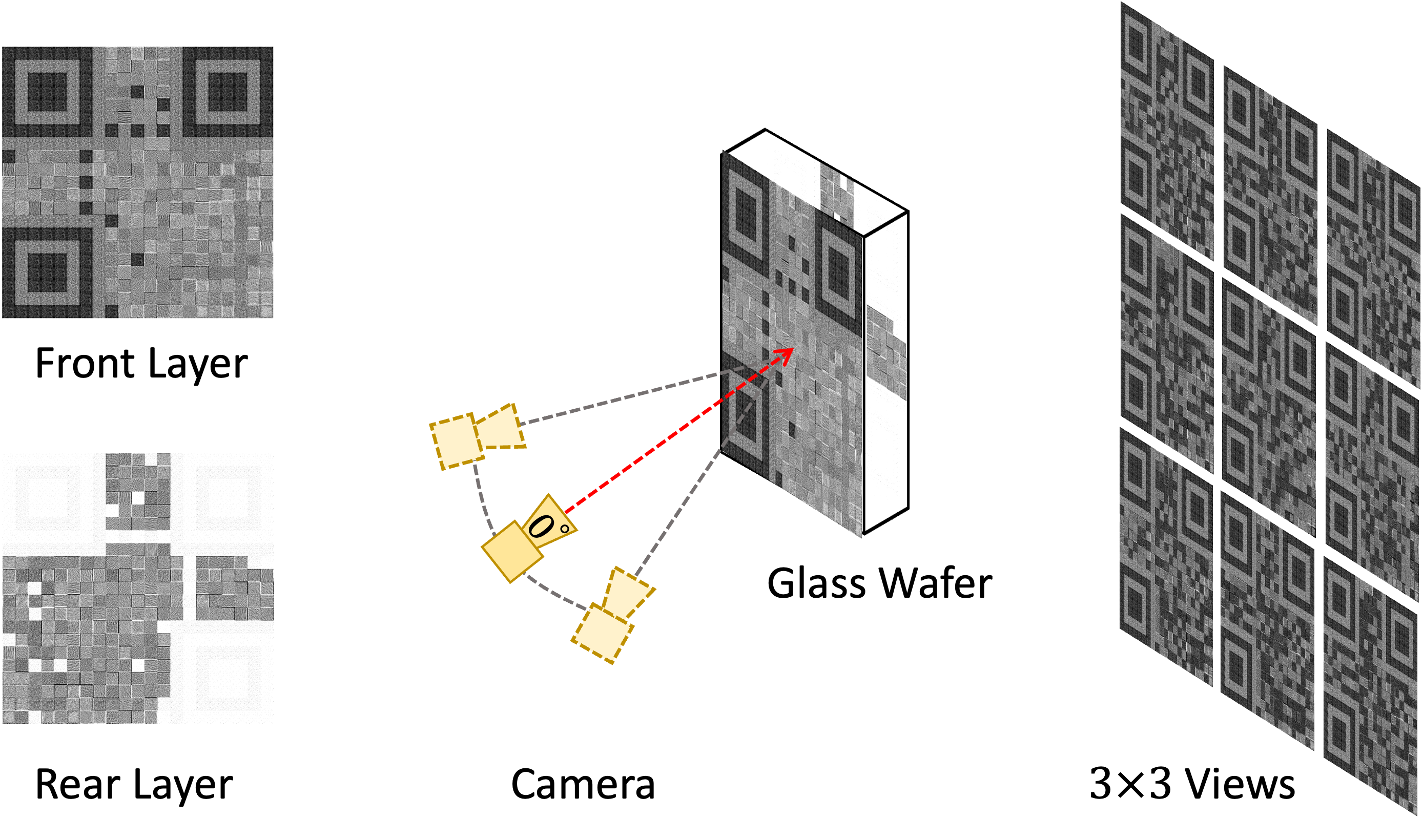}
	\caption{\emph{Angular measurement via QR-design patterns.} We
          encode angular information into the optimized QR-design patterns generated by
          binary structures printed on the two sides of a
          glass wafer. Angular shifts create different superposed QR codes that can be scanned in the low frequency. Each viewing direction from the $3 \times 3$ views on the right can perform accurate angular measurements.}

\label{fig:teaser}
\end{figure}

\section{Introduction}
\label{intro}
We introduce an optical marker designed for precise measurement of angular and directional information in a discrete manner. This carefully designed optical element, known as QR-Tag, is founded on the parallax effect and QR code principles, enabling the measurement of the subtle movement of an object with easily readable angular values.  Minute adjustments in the relative offsets of two high-resolution binary structures produce distinct superimposed QR codes, detectable by a camera at low frequencies. In the two-layer design of QR-Tag, these offsets correspond to the parallax under different observation directions, resulting in apparent visual shifts in the superimposed patterns along various directions. We have developed a comprehensive image formation model for generating two high-resolution binary layers by applying binary matrix factorization.

Several prior works have employed the two-layer design, parallax effect, and \m effect to continuously measure directional information. Examples include camera tracking for 3-DoF translations~\cite{banks2019use,xiao2021moireboard}, and camera pose estimation for 6-DoF extrinsic parameters~\cite{ning2022moirepose}, all of which leverage the magnification properties of Moiré patterns. More recently, the \mt~\cite{qiu2023moiretag} exploits the \m effect to encode the high accuracy angular information into a \m pattern, enabling precise object angular measurement and tracking. The \mt is capable of continuously tracking the angular shift at low frequency with high accuracy. Nonetheless, it does not yet encompass a complete QR code design, which remains the primary focus of this study. We propose designing two-layer binary patterns that encode high-frequency binary information, while the camera captures the shifted geometric alignment of superimposed patterns.These patterns, when observed at low frequencies, become grayscale QR codes that can be scanned from various angles, effectively representing the actual angular information. To the best of our knowledge, this is the first work that comprehensively utilizes QR-design and the parallax effect to achieve high-precision object angular measurements in discrete fashion. Furthermore, we delve into expansions of the core methodology for estimating positional information based on a single snapshot image. Specifically, we make the following contributions:

\begin{itemize}

\item We present a novel, cost-effective, reliable, computationally efficient, and highly accurate method for object tracking. This method can track an object at discrete angles without requiring camera calibration.

\item We utilize the parallax effect and QR code design to construct a two-layer marker for measuring the subtle motion of an object, offering easily readable angular information. The two-layer binary QR-design is optimized through binary matrix factorization.

\item A comprehensive image formation model with pixel-wise design can encode a variety of readable angular information into the two-layer optical marker.

\item We use two simulation results, with $3 \times 3$ views and $5 \times 5$ views, to demonstrate the utility of the proposed QR-design optical marker for measuring precise directional information.

\end{itemize}

\section{Related Work}
\label{related}

\paragraph{Optical tracking:}
Conventional Checker boards~\cite{zhang2000flexible} or fiducial markers~\cite{garrido2014automatic,fiala2005artag,benligiray2019stag,wang2016apriltag,degol2017chromatag} are commonly employed in various scenarios, including robotics~\cite{song2017cad}, visual odometry~\cite{harmat2015multi}, and augmented/virtual reality systems. While these approaches excel in accurately determining object positions, they often struggle with providing reliable rotational or orientation information. The accuracy of this orientation data heavily relies on meticulous camera intrinsics calibration and the effective distribution of fiducial markers across the entire field of view. These characteristics pose several challenges. Firstly, the requirement for calibration complicates real-world applications outside controlled environments, as it can vary with focus or zoom settings. Secondly, achieving comprehensive field-of-view coverage may limit the ability to track small objects at greater distances. Other optical methods for angular measurements exist, such as autocollimators~\cite{geckeler2012novel,chen2017optical}, which offer superior sensitivity and precision. However, they are even more challenging to deploy in uncontrolled settings.

\paragraph{\m design tracking:}
There are a plenty of methods that utilize the \m effect's amplification property for measuring the directional information of an object. Hideyuki et al.~\cite{Tanaka2012ApplicationOM,tanaka2014solution} propose to apply \m patterns for solving orientation accuracy in frontal observation and pose ambiguity. The Inogon~\cite{ship} serves as a pioneering instance of employing the \m effect for angular measurement. It functions as a passive visual indicator to guide ships through narrow shipping lanes. Banks et al.~\cite{banks2019use} propose to apply a two-layer line \m design for 3D position estimation. Nevertheless, this approach has some limitations. The working volume is restricted, the \m patterns are considerably larger than the working volume, and the camera must maintain a fronto-parallel orientation to the two planes. Subsequently, some of these constraints were eased, as demonstrated in the work by Xiao et al.~\cite{xiao2021moireboard} and Jo et al.~\cite{jo2023moire}, permitting tilted camera angles. However, the work from Jo et al. requires camera pre-calibration. If changing to another camera, the measurement requires camera calibration again, and the accuracy depends on the size of the target. Ning et al.~\cite{ning2022moirepose} introduce a novel approach for 6 DoF pose estimation using \m patterns generated by re-photographing a monitor with a digital camera. However, it's important to note that this method necessitates precise calibration tailored to the unique camera and display pairing. In very recent work, Qiu et al.~\cite{qiu2023moiretag} offers a novel thin markers with a small volume that works with a large standoff distance and primarily targets angular measurements with additional camera intrinsic measurement. 

In comparison to these approaches, our method offers easily readable passive markers that are effective at longer standoff distances, primarily focusing on discrete angular measurements. Our two-layer design applies the parallax concept, which has also been employed in compressive displays. Similarly, these methods encode high-dimensional information into the design multiplicatively (e.g., Lanman et al. 2011; Wetzstein et al. 2012; Heide et al. 2014a; Peng et al. 2017)~\cite{lanman2011polarization,wetzstein2012tensor,heide2014cascaded,peng2017mix}.

\section{Overview}
\begin{figure}[ht]
	\centering
	\includegraphics[width=1\linewidth]{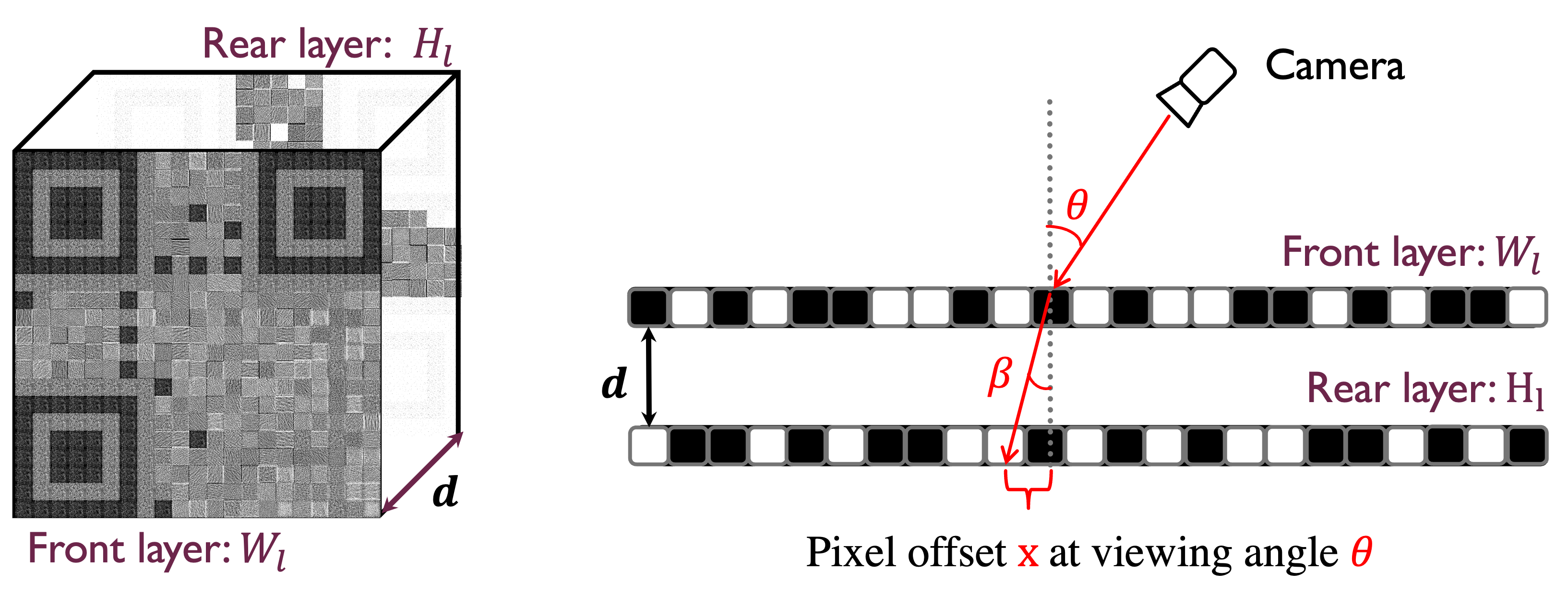}
	\caption[Mapping between angles and spatial offsets]{Left: We display a double-sided glass plate encoded with binary QR-design. Each layer contains three finder patterns for captured image rectification. Both layers consist of binary patterns composed of 0s and 1s. Right: By leveraging the geometric alignment of these two layers, we can establish the relationship between pixel offset and viewing angle $\theta$.}
	\label{fig:pp1}
\end{figure}

Our goal is to present a novel optical-based visual marker, QR-Tag, designed for object angular measurement and tracking. QR-Tag operates on the principles of the parallax effect and QR code design, as illustrated in Fig.~\ref{fig:pp1}. By multiplying two high-resolution binary patterns with different binary combinations, we generate superimposed patterns that can be captured by a phone camera at a low frequency, resulting in grayscale images. These grayscale QR codes encode the relative offset, representing the angular information. 
To achieve the most accurate and rapidly readable camera angular estimation results, we have developed an algorithm based on binary matrix factorization to optimize two binary layers. These optimized layers are then fabricated on both the front and back sides of a glass wafer with a set thickness.

\subsection{Mapping Between Angles and Spatial Offsets}
We utilize the parallax generated from the geometric alignment of the two-layer marker to measure the camera position. The physical realization of QR-Tag is to print the two-layer designs onto the two sides of a thin glass wafer. Therefore, we need to account for refraction on the glass-air interface when mapping the captured QR-Tag to physical angles. In our design, the glass wafer has a thickness of $\vd=510\mu m$, the glass’s refractive index is $\nb \approx 1.4600$, and the refractive index of air is $\na \approx 1$. Considering Snell’s law~\cite{shirley1951early}, $\na \sin \theta = \nb \sin \beta$. The relationship between the camera viewing angle $\theta$ and pixel offset $x$ relative to the superposed two QR binary layers is $\vx = \vd \tan(\arcsin(\frac{\sin \theta}{\nb}))$.

\section{Computational Design}
In this section, we derive a comprehensive image formation model for optimizing and encoding binary patterns on the two layers of the marker based on QR-design. We have previously established a straightforward mapping between angles and spatial offsets. In the following, for the sake of notation simplicity, we opt to base all derivations on spatial offsets.

\subsection{Image Formation Model}
The image formation model is based on a two-layer binary design, incorporating optimized black-and-transparent patterns. We aim to observe different readable QR codes as changing the viewing angles relative to the designed QR-Tag. Fig.~\ref{fig:teaser} depicts nine viewing angles as an example. 


The designed two-layer target on the left contains high-resolution binary patterns. However, the high-frequency patterns are either suppressed due to limited camera resolution, for instance, at large standoff distances from the target, or they can be filtered out after capture. Consequently, the camera-captured images provide low-resolution information. Based on this phenomenon, we present our image formation model as follows:

\begin{equation}
\label{eq:(pp3)} 
V = B(WH).
\end{equation}

where $V$ represents the known low-resolution input target, which is a grayscale image encoding angular information. As we know, a standard QR code exhibits intensity contrast with values of 0 and 1. We create the input QR target by mapping the 0 intensity value to 0.2 and the 1 value to 0.5. $WH$ is the outer product of two high-resolution rank-one vectors, The two-layer high-resolution binary patterns are reshaped into two rank-one vectors $W$ and $H$. We then apply a downsampling matrix $B$ to the high-resolution matrix $WH$. Here, we aim to minimize the difference between the binary matrix $B(WH)$ (composed of 0s and 1s) and the grayscale input target $V$ (composed of 0.2s and 0.5s).
 
\begin{figure}[ht]
	\centering
	\includegraphics[width=.9\linewidth]{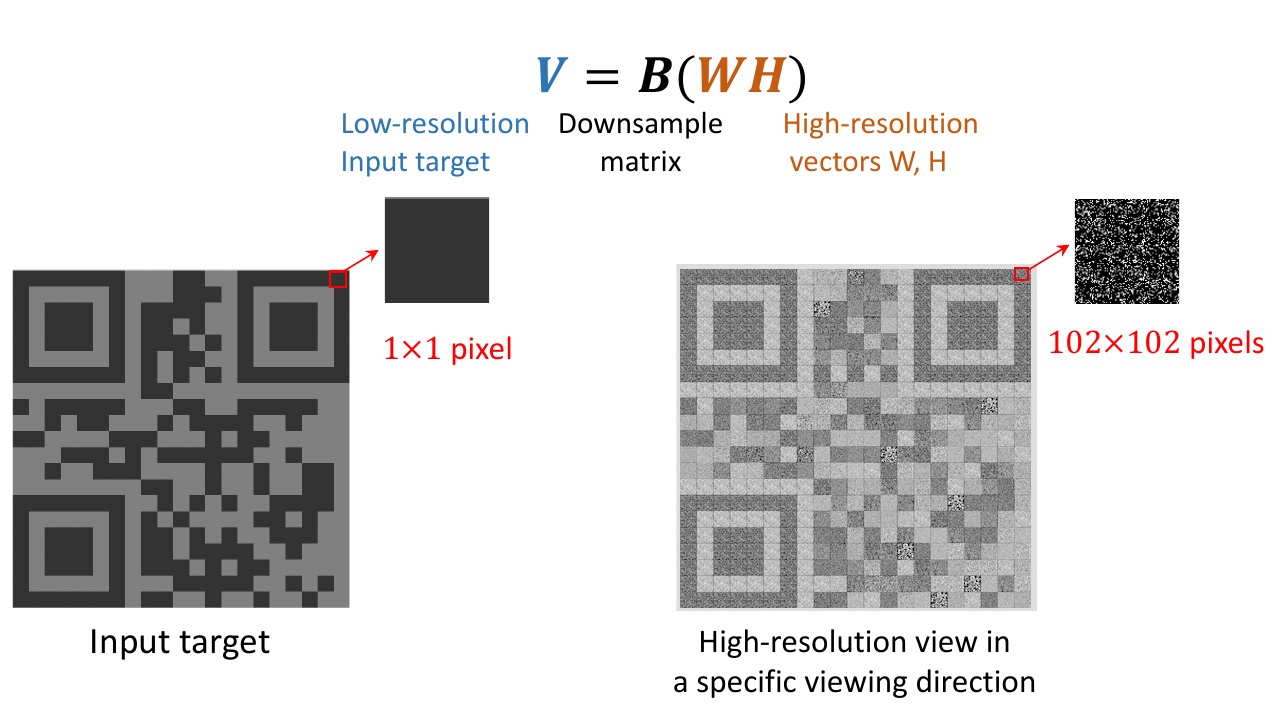}
	\caption[Image formation model]{Image formation model. We show the low-resolution QR-based input target alongside the superposition of the optimized high-resolution view. The intensity contrast of the low-resolution input target is from 0.2 to 0.5. The red corner on the left image consists of a single $1 \times 1$ pixel, while the corner on the right image comprises $102 \times 102$ pixels (composed of 0s and 1s). It's evident that the downsampling scale here is 100, with an additional $2 \times 2$ pixels from the high-resolution image forming the added boundary.}
	\label{fig:low and high}
\end{figure} 

In Figure \ref{fig:low and high}, we provide a single-view example where we encode high-frequency information into two optimized binary layers, while the camera captures and scans the superimposed grayscale QR code in the low frequency. In our design, instead of focusing on a single viewing direction, both the low-resolution matrix and the outer product matrix encode information from multiple viewing angles.

\subsection{Multi-view Design}
The previous section introduced the image formation model and the concept of single-view reconstruction. However, we are not satisfied with creating only one viewing direction with a double-layer design. We propose optimizing two layers so that each viewing direction of the superposed pattern displays a different QR-Code, which can be scanned and read at a specific viewing angle. 

\begin{figure}[ht]
	\centering
	\includegraphics[width=1\linewidth]{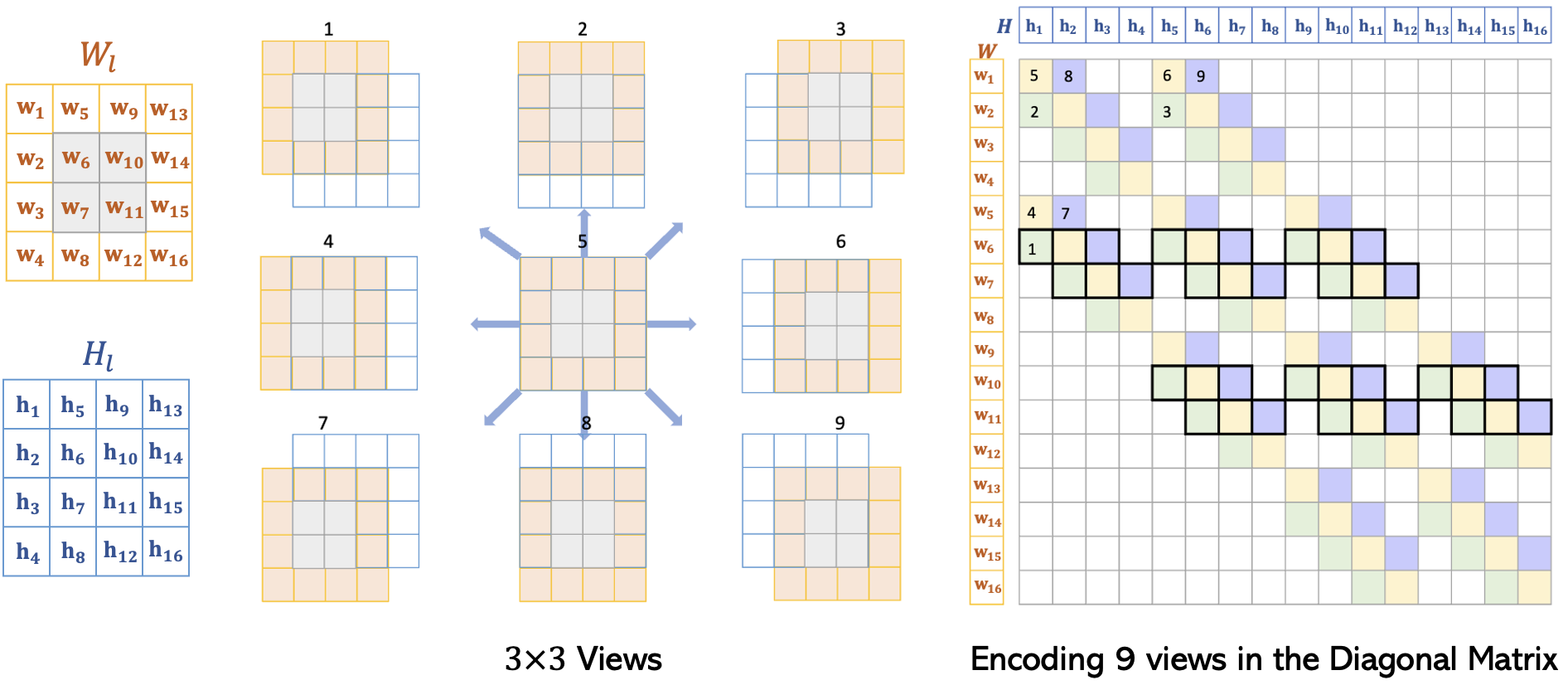}
	\caption[$3 \times 3$ viewing directions]{Left: two-layer design. Middle: $3 \times 3$ viewing directions are derived from two high-resolution binary layers ($W_l$ and $H_l$), with each grid representing a pixel relative to the low-resolution input. Right: We show the permutation of the encoded viewing angles.}
	\label{fig:pp3}
\end{figure}

We illustrate this concept using a $3 \times 3$ views example in Fig.~\ref{fig:pp3}. The two layers $W_l$ and $H_l$ on the left represent the high-resolution binary layers, reshaped from $W$ and $H^T$. The gray area in the center of $W_l$ contains the information, while the additional orange area serves as periodic padding to avoid boundary issues when changing the viewing angles. In the middle, the image $5$ representing the front view of the two-layer design. When shifting the viewing position horizontally from the center towards the right by a specific offset, image $6$ becomes visually observable. Each pixel offset corresponds to a specific angle, which can be determined through the mapping between the pixel offset and the viewing angle. The images on the right displays the outer product of $W$ and $H$, reshaped from $W_l$ and $H_l$. The nine views are encoded along the diagonal directions, with each marked by numbers at the first pixel. Additionally, pixels with black boundaries represent the actual angular information within the image. As a result, we can extract nine diagonals from the optimized diagonal matrix outer product. These scannable grayscale QR codes representing the viewing directions. To achieve this, we need to first solve a binary matrix factorization problem. 

\subsection{Pixel-Wise Image Formation}
\begin{figure}[h]
	\centering
	\vspace{-10pt}
	\includegraphics[width=1\linewidth]{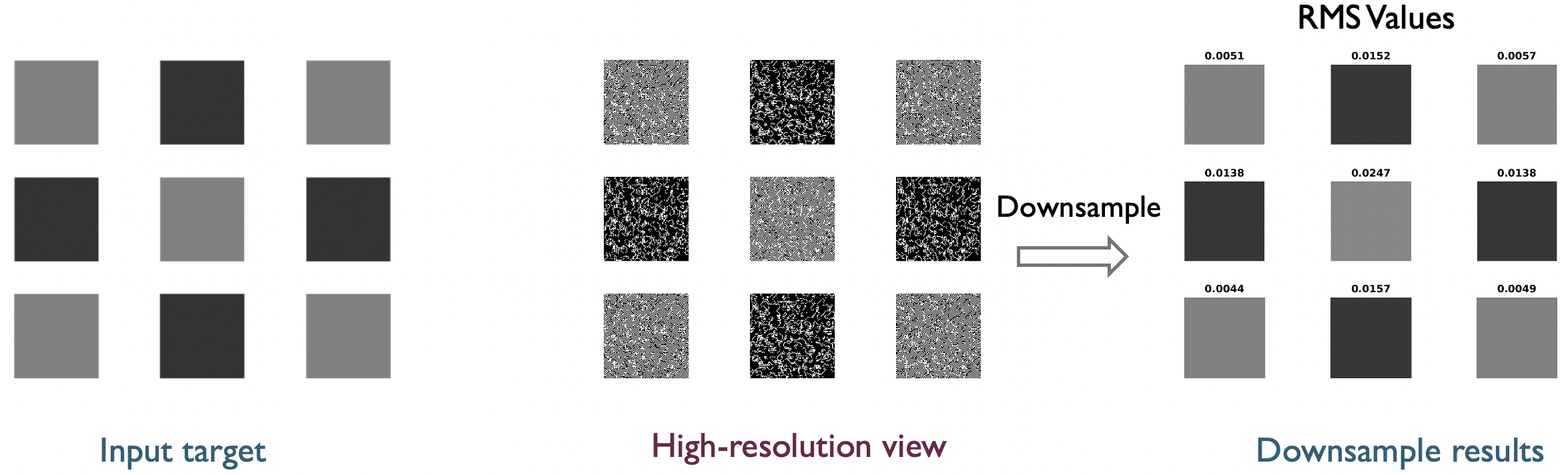}
	\caption[Pixel-wise image formation]{\emph{Pixel-wise image formation.} Left: Nine grayscale low-resolution pixels from each viewing direction. Middle: Nine binary high-resolution views. Right: The RMS errors between the input targets and the downsampled results.}
	\label{fig:pp5}
\end{figure}

To ensure the versatility of the marker, we employ a pixel-wise design capable of reconstructing QR code patterns encoding any text information. This approach can thus be adapted to track various types of angular information. Taking the $3 \times 3$ views case as an example, the low-resolution input target comprises 9 pixels, each of which can be either 0.2 or 0.5. This results in a total of $2^9 = 512$ different combinations (see Fig.~\ref{fig:pp5}).

\begin{figure}[ht]
	\centering
	\includegraphics[width=1\linewidth]{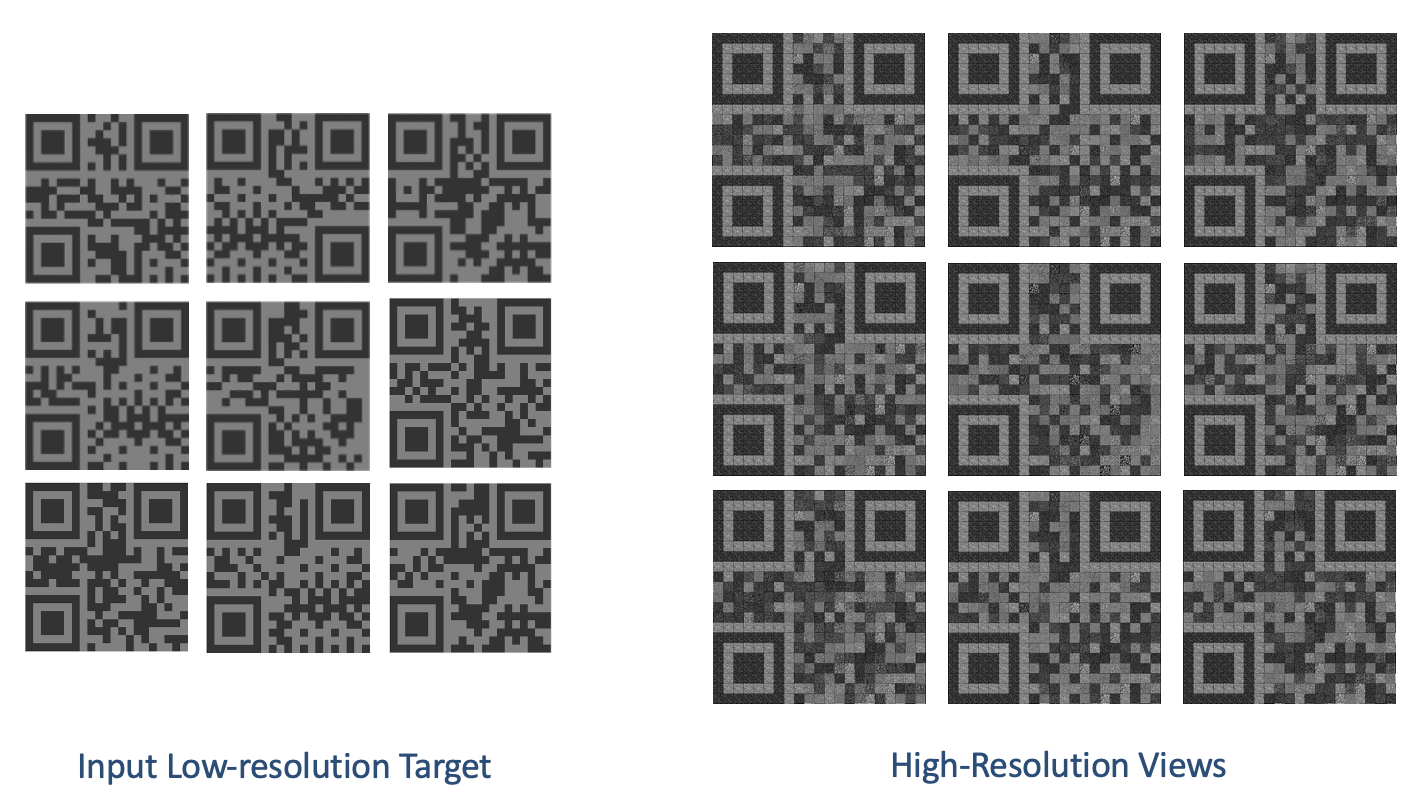}
	\caption[Group pixel-wise images.]{We aggregate the optimized pixel-wise results to formulate the high-resolution two-layer design.}
	\label{fig:pp6}
\end{figure}

After solving the optimization problems for all combinations, we can design the input target that contains various information by aggregating the pixels. Fig.~\ref{fig:pp6} illustrates the grouped outcomes of nine high-resolution views. Each viewing direction is readable through a phone camera, emphasizing that pixel-wise image formation does not impact the final results. In fact, this approach enhances the efficiency of our designed algorithms.

\section{Inverse Problem}
With the image formation model in Eq.~\ref{eq:(pp3)}, we formulate the following constrained optimization problem:
\begin{equation}
\begin{aligned}
\min_{W \in \mathbb{R}^{m \times r}, H \in \mathbb{R}^{r \times n}} \quad & \frac{1}{2} \parallel C \circ (V - B(W H)) \parallel^{2}_{F} + \lambda_1 \parallel W \parallel^{2}_{2} + \lambda_2 \parallel H \parallel^{2}_{2} \\
\textrm{s.t.} \quad & W, H \in {0,1}\\
\end{aligned}
\end{equation}
where $W\in \mathbb{R}^{m \times r}$ and $H^T \in \mathbb{R}^{n \times r}$ represent two rank-one vectors and $C$ is the coefficient. $\lambda_1$ and $\lambda_2$ are the coefficients of the $L_2$ regularizers. We aim to optimize the above function to discover the optimal combination of the two layers. Algorithm 1 shows pixel-wise optimization using weighted non-negative matrix factorization~\cite{ho2008nonnegative}, including Hadamard (element-wise) matrix division.

\begin{algorithm}
\caption{Weighted Rank-1 Nonnegative Matrix Factorization (WNMF)}
\begin{algorithmic}
\STATE \textbf{Initialization} $w^0 = w_{\text{rand}}$ and $h^0 = h_{\text{rand}}$
\WHILE {not converge}
\STATE $w =$ \textbf{periodic padding} $(w^0)$
\STATE $w_{k+1} \leftarrow w_k \circ \frac{B^T(C \circ V)h^T_k}{B^T(C \circ [B(w_kh_k)])h^T_{k}+2 \lambda_{1}w_k} $
\STATE $h_{k+1} \leftarrow h_k \circ \frac{w^{T}_k B^T(C \circ V)}{w^T_k B^T (C \circ [B(w_k h_k)]) + 2 \lambda_{2}h_k}$
\ENDWHILE  
\end{algorithmic}
\end{algorithm}

We apply Algorithm 2 to solve the binary matrix factorization~\cite{zhang2007binary} problem, specifically, we employ the Sigmoid function to identify the optimal pair of thresholds, $w_{thresh}$ and $h_{thresh}$, for generating binary $w$ and $h$. In particular, the Sigmoid function is represented as $sigmoid(x, y)$ in Algorithm 2. We iteratively adjust the thresholds, $a_n$, in $\phi(x) = \frac{1}{1+e^{-a_n(w-w_{\text{thresh}})}}$, shaping the Sigmoid function $\phi(x)$ from gentle to steep in each iteration. We search for the thresholds, $w_{\text{thresh}}$ and $h_{\text{thresh}}$, to obtain two binary layers.

\begin{algorithm}
\caption{Weighted Rank-1 Binary Matrix Factorization (BMF)}
\begin{algorithmic}
\STATE \textbf{Initialized} $w^0$ and $h^0$ \textbf{from Algorithm 1, and initialize} $w^0_{\text{thresh}}$, $h^0_{\text{thresh}}$, $a^0$
\FOR {$a = a_{k}$ \textbf{to} $a_{K}$}
\WHILE {not converged}
\STATE $[w^k_{\text{thresh}}, h^k_{\text{thresh}}] =$ \textbf{gradient descent} $(C, V, w^0, h^0, a^0, w^0_{\text{thresh}}, h^0_{\text{thresh}})$
\ENDWHILE
\STATE $w_{k} = \textbf{sigmoid}(a_{k}, w_{k-1}-w^k_{\text{thresh}})$
\STATE $h_{k} = \textbf{sigmoid}(a_{k}, h_{k-1}-h^k_{\text{thresh}})$
\ENDFOR
\STATE $w = \textbf{float2binary}(w_k, w^k_{\text{thresh}})$
\STATE $h = \textbf{float2binary}(h_k, h^k_{\text{thresh}})$
\end{algorithmic}
\end{algorithm}


\section{Simulation Results}
\begin{figure}
	\centering
	\includegraphics[width=1\linewidth]{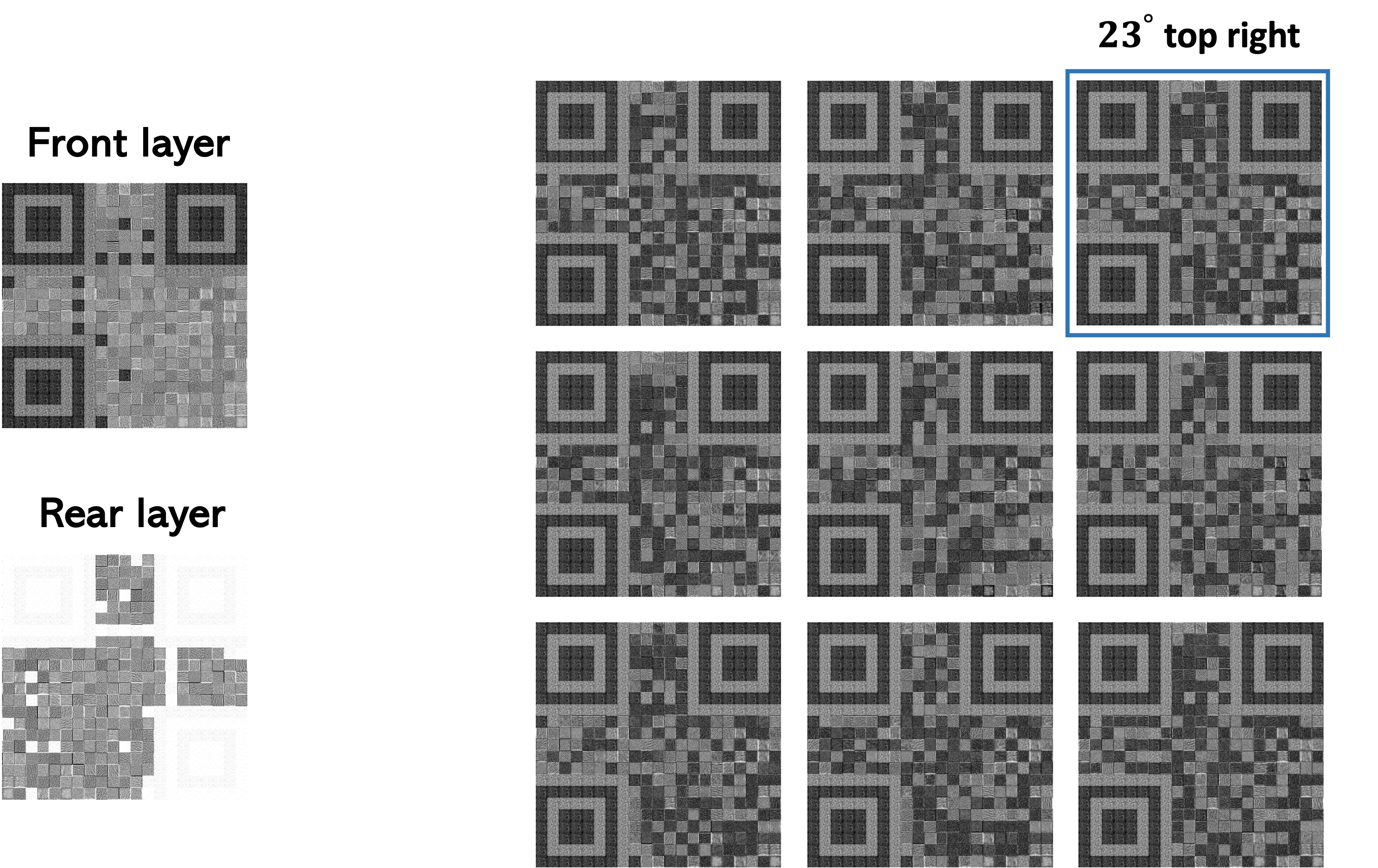}
	\caption[$3 \times 3$ viewing directions]{$3 \times 3$ views. On the left, we display the optimized front layer and the rear layer. Both layers are comprised of 0s and 1s. On the right, we show the scannable views relative to each viewing direction.}
	\label{fig:s1}
\end{figure}

\begin{figure}
	\centering
	\includegraphics[width=1\linewidth]{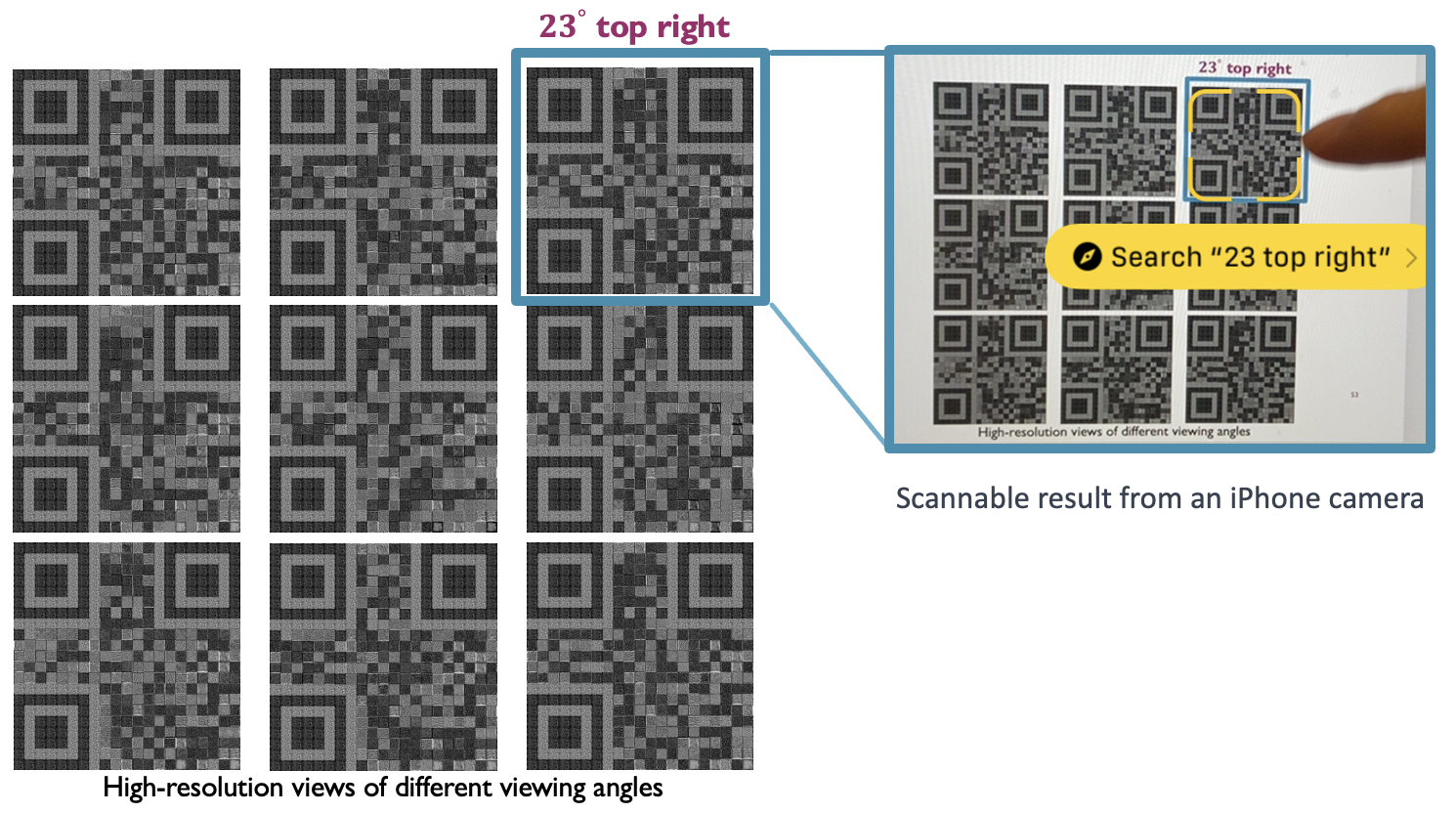}
	\caption[$3 \times 3$ viewing directions]{$3 \times 3$ views. We use an iPhone camera to scan the QR codes from the reconstructed viewing angles. The result in the top-right corner represents a viewing angle of $23^\circ$ degrees relative to the image in the middle ($0^\circ$).}
	\label{fig:s2}
\end{figure}

One of the simulation results in Fig.~\ref{fig:s1} contain $3 \times 3$ viewing angles, which are $23^\circ$ in 9 discrete directions. The other simulation results in Fig.~\ref{fig:s2} comprise $5 \times 5$ views capable of achieving 25 discrete viewing angles.Based on the simulation results, we have successfully encoded directional information into the high-resolution two-layer design, enabling the generation of scannable QR codes from different viewing directions.

\begin{figure}
	\centering
	\includegraphics[width=1\linewidth]{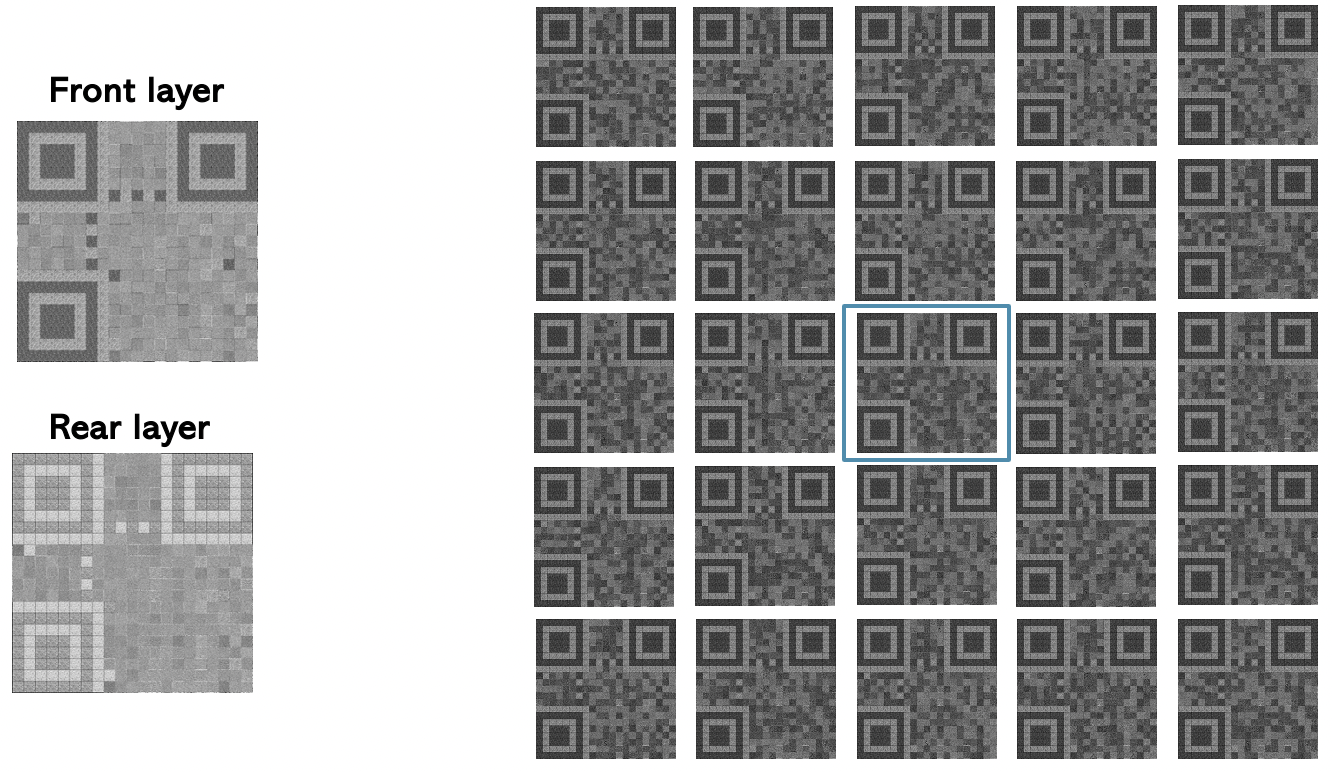}
	\caption[$5 \times 5$ viewing directions]{$5 \times 5$ views. On the left, we display the optimized front layer and the rear layer. Both layers are comprised of 0s and 1s. On the right, we show the scannable views relative to each viewing direction.}
	\label{fig:pp8}
\end{figure}

\section{Prototype}
To create a compact measuring instrument, we produced a prototype QR-Tag on both sides of a 4-inch fused silica wafer using cutting-edge photolithography techniques. The mask patterns were etched into chromium layers to achieve optimal absorption. However, as chromium is highly reflective, we initially deposited a layer of $100~nm$ of silicon dioxide ($SiO_{2}$) on both sides of the wafer using plasma-enhanced chemical vapor deposition (PECD). The $SiO_{2}$ film served as a light absorber to minimize interreflection between the chromium layers. We utilized the front-to-back alignment technique provided by the contact aligner ($EVG6200\infty$). 

\begin{figure}
	\centering
	\includegraphics[width=.7\linewidth]{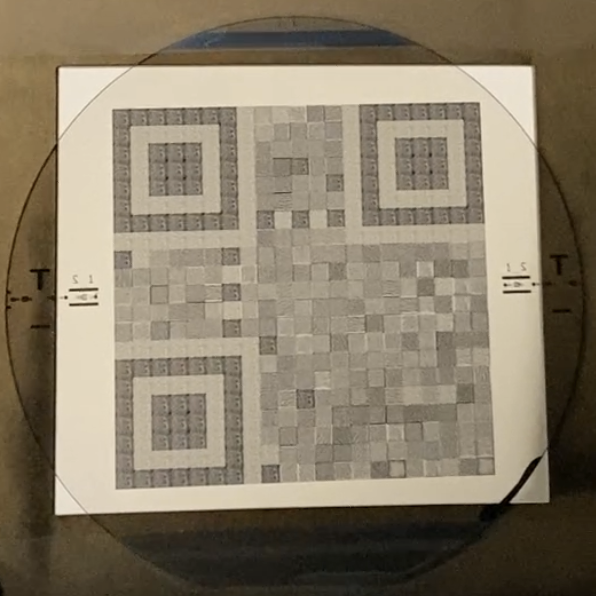}
	\caption[Prototype]{ $3 \times 3$ views prototype.}
	\label{fig:pp9}
\end{figure} 

\section{Discussion}
We present a vision-based optical visual marker, the QR-Tag, which incorporates the high-resolution binary patterns to achieve object angular measurement. Our approach involves encoding viewing directions into a two-layer high-resolution QR-based target, while the camera captures and scans the observed low-resolution images for precise angular measurement. The QR-Tag design is scannable and can cover large capturing distances, accommodating a wide range of viewing angles with high sensitivity. Furthermore, the compact and versatile nature of the design meets the requirements of various measuring scenarios, including virtual reality (VR) and augmented reality (AR), autonomous vehicle, aerial vehicle, and robotics. What sets our 3-DoF object tracking design apart is its ability to provide a snapshot procedure without the need for calibration or complex environmental control, unlike other state-of-the-art methods.

While the simulation results are indeed fascinating, as illustrated in the prototype in Fig.~\ref{fig:pp9}, we encountered significant challenges when attempting to scan and read the QR codes. We fabricated over ten prototypes with various designs and materials, yet none of them are readable by smart phone camera. There are several reasons causing this effect.

\begin{itemize}

\item As we explained in the prototype fabrication, we used the chromium material for pattern fabrication. Unfortunately, the deposition method of this material by PECVD on a thin film on a glass wafer makes the layer very reflective. This creates significant inner reflections between the top and bottom layers of the prototype and makes scanning the code complex. We have tried to use two layers of SiO2/Cr absorber like in this paper~\cite{kim2020high}, but unfortunately, the pattern was very dark and difficult to scan. We also tried Cooper Dioxyde, but unfortunately, we could not reduce reflection.

\item Despite the high precision in fabrication accuracy and the alignment of the two layers (approximately 1 micrometer), strong interreflection within the glass wafer makes the design challenging to scan.

\item The intensity contrast of the input grayscale image is 0.2 and 0.5, which are determined through multiple experiments in binary matrix factorization. The low intensity contrast is another factor that affects the final results.

\end{itemize}

In the future, the improvement in the fabrication stage can be done by using dark material and changing the silica wafer into a less reflective material. We hope our proof of QR-Tag can motivate further research for high-precision angular measurement.

\newpage
\bibliographystyle{splncs}
\bibliography{biblio}

\end{document}